\documentclass[lettersize,journal]{IEEEtran}
\usepackage{amsmath,amsfonts}
\usepackage{algorithmic}
\usepackage{algorithm}
\usepackage{array}
\usepackage[caption=false,font=normalsize,labelfont=sf,textfont=sf]{subfig}
\usepackage{textcomp}
\usepackage{stfloats}
\usepackage{url}
\usepackage{verbatim}
\usepackage{graphicx}
\usepackage{cite}
\usepackage{xcolor}
\usepackage{booktabs}
\usepackage{pifont}
\usepackage{multirow}
\usepackage{tikz}
\usetikzlibrary{shapes.geometric, arrows, positioning, calc}
\usepackage{framed}
\usepackage{eurosym}
\usepackage{comment}

\begin{document}

\title{MxGPS: Multiplex Graph Transformers \\ for a Power Grid Foundation Model}

\author{
    \IEEEauthorblockN{Charilaos Papaioannou, Ioannis Tsantilas, Dimitris Giannakakos, Vasilis Michalakopoulos, Sotiris Pelekis, Vangelis Marinakis, Arsam Aryandoust, Antonello Monti, Ricardo J. Bessa, Pedro P. Vergara, Jochen Cremer, Elissaios Sarmas}
}




\maketitle

\begin{abstract}

Single-task fine-tuning of graph neural networks~(GNNs) for power grid problems exhibits a systematic failure mode: models that achieve the lowest in-distribution error degrade the most under topology shift.
We term this \emph{topology overfitting}: the tendency of task-specific gradient signals to encode relational structure particular to the training topologies rather than the underlying physics, causing models to fail on unseen grids despite strong in-distribution performance.
To expose and address this failure mode, we introduce MxGPS~(Multiplex GPS), a multiplex graph transformer that runs $K$ task-specialised GPS branches over a shared node encoder, jointly trained on Static State Estimation~(SSE) and AC Power Flow~(PF) via a self-supervised pre-training and multi-task fine-tuning protocol, with a cross-branch attention module evaluated in ablation.
The joint SSE$+$PF objective forces the shared encoder to simultaneously satisfy complementary gradient signals, preventing it from overfitting to topology-specific relational structure.
Under a 3-fold sliding-window cross-validation spanning four unseen topologies (14-, 24-, 162-, and 300-bus), MxGPS attains $0\%$ boundary violation rate~(BVR) on all four zero-shot Power Flow topologies.
Critically, models with substantially lower in-distribution PF error degrade by $190\%$--$1400\%$ under topology shift, whereas MxGPS degrades by only $39\%$, an inversion that directly implicates topology overfitting as the failure mechanism rather than insufficient model capacity.
With only 1.6M parameters ($12\times$ fewer than the GridFM reference baseline), MxGPS demonstrates that multi-task joint training is a principled and parameter-efficient mechanism for topology-agnostic generalisation in power grid foundation models.

\end{abstract}


\IEEEaftertitletext{\vspace{-2\baselineskip}}

\section{Introduction} \label{sec:introduction}

\IEEEPARstart{T}{he} fundamental objective of power grid operation is the reliable, secure, and cost-effective delivery of energy. Traditionally, the core computational challenges of grid operation, namely AC power flow~(PF), static state estimation~(SSE), optimal power flow~(OPF), and N-$k$ contingency analysis, are addressed by iterative numerical solvers such as Newton-Raphson and interior-point methods \cite{stott1974review}. However, the high penetration of intermittent renewables and bidirectional flexible loads has introduced unprecedented uncertainty and multi-directional information flow \cite{liao2021review}. While these solvers are accurate, they are computationally intensive, topology-specific, and ill-suited to the rapid situational changes that characterize modern grid operation.

Machine learning approaches, and graph neural networks~(GNNs) in particular, have emerged as a promising complement to traditional solvers by learning surrogate models that generalise across operating conditions~\cite{ghamizi2024safepowergraph}. GNNs are naturally suited to power grid networks: buses are nodes carrying electrical state variables, transmission lines are edges carrying admittance parameters, and the physics of the grid is expressed as algebraic constraints on these graph-structured quantities~\cite{zhou2020graph}. Despite this alignment, existing GNN approaches for power grids train separate models per task~\cite{zhao2020deepopf,liao2021review}, precluding representation sharing across operational objectives and failing to exploit the physical consistency constraints shared across SSE, PF, and contingency analysis.

GridFM~\cite{hamann2024foundation} takes a step toward a unified model by pre-training a General, Powerful, Scalable (GPS)-based graph transformer~\cite{rampasek2022recipe} on the Masked Grid Task~(MGT): a self-supervised masked feature reconstruction objective analogous to masked autoencoders in vision~\cite{he2022masked} and BERT in language~\cite{devlin2019bert}. GridFM demonstrates that a single model pre-trained on a diverse grid portfolio can be fine-tuned with minimal labelled data for PF and SSE, transferring across topologies. However, GridFM remains a fundamentally single-task architecture: it learns one reconstruction objective with one set of edge attention patterns, conflating the distinct relational structures that PF (requiring admittance-aligned coupling) and SSE (requiring measurement-noise-robust structure) impose on the same physical topology. We argue this conflation is a structural limitation, not merely a scaling issue: SSE should up-weight reliable measurement buses and down-weight noisy ones, while PF should align edge attention with the admittance matrix. No single graph representation can optimally serve both objectives simultaneously, a limitation noted in the GridFM paper itself~\cite{hamann2024foundation}.

We propose \textbf{MxGPS}~(Multiplex GPS), an architecture that makes task-specific relational specialisation explicit by running $K$ parallel GPS branches over the same physical node set, each steered toward a different operational task by its own gradient signal. A shared node encoder provides a common physical representation before branching; a cross-branch attention module (evaluated in ablation as MxGPS-cross) enables the task experts to mutually regularise each other during training, analogous to cross-head attention in multi-head Transformers~\cite{vaswani2017attention}. An AC power flow consistency loss, derived from the Bus Injection Model~(BIM), is applied across all branches simultaneously as a shared physics-informed self-supervised signal. Critically, this joint multi-task objective acts as a structural mechanism for \emph{topology-agnostic} generalisation: forced to simultaneously satisfy gradient signals from both SSE and PF, the shared encoder cannot overfit its representations to any particular training topology's relational structure, yielding embeddings that transfer substantially better to unseen grids than those learned by single-task fine-tuning.

\paragraph{GNNs for power grids}
GNNs have been applied to power flow, state estimation, optimal power flow~\cite{zhao2020deepopf}, and security assessment~\cite{ghamizi2024safepowergraph,zhang2025moe}, approximating traditional solver outputs at a fraction of the computational cost, but typically in single-task, single-topology settings.

\paragraph{Graph transformers and self-supervised learning on graphs}
The Transformer architecture~\cite{vaswani2017attention} has been extended to graphs through several lines of work~\cite{muller2023attending}. GraphGPS~\cite{rampasek2022recipe} combines a local message-passing network (GatedGCN; \cite{bresson2017residual}) with a global Transformer layer, achieving strong performance across benchmarks~\cite{dwivedi2023benchmarking}; its structural encodings, Random Walk Structural Encoding~(RWSE) and Laplacian Positional Encoding~(LapPE)~\cite{dwivedi2022graph}, capture topological position, critical for power grids where bus roles are tied to graph-theoretic properties. Heterogeneous graph transformers~\cite{hu2020heterogeneous} handle multi-type node and edge relations, relevant for grids with bus and generator nodes. On the self-supervised side, masked reconstruction pre-training~\cite{he2022masked,devlin2019bert} yields transferable representations; GraphMAE~\cite{hou2022graphmae} extends it to graphs via the Scaled Cosine Error~(SCE) loss, which GridFM adapts for the MGT objective in place of Mean Squared Error~(MSE). GridFM~\cite{hamann2024foundation} adopts GPS as its backbone, making this architecture the natural starting point for our work.

\paragraph{Multi-task learning and multiplex networks}
Shared representations across tasks provide implicit regularisation and improve generalisation~\cite{caruana1997multitask}. MTGC3~\cite{qin2023multitask} jointly optimises task collaboration and difficulty scheduling in multi-task GNN architecture search, while Graph Mixture of Experts~(MoE)~\cite{wang2024graphmoe} explicitly models expert diversity to prevent representation collapse. A multiplex network represents the same node set across multiple relational layers, each capturing a different interaction type; cross-layer attention~\cite{sharma2026mind} has been studied for link prediction in such networks. In MxGPS, each branch is a relational layer specialised for one operational task, paralleling the expert decomposition of MoE, and cross-branch attention is the analogue of cross-layer interaction.

\paragraph{Physics-informed learning and foundation models}
Physics-informed neural networks~\cite{karniadakis2021physics} enforce known physical laws as soft constraints during training; in power grids, the AC power balance equations~(Eqs.~\ref{eq:pbal}--\ref{eq:qbal}) of the BIM are the natural constraints. The success of foundation models in language~\cite{brown2020language} and vision has motivated analogous efforts in scientific domains: GridFM is the first foundation model explicitly designed for power grids, pre-trained on a diverse grid portfolio generated by \texttt{gridfm-datakit}~\cite{puech2025gridfm} with a BIM-based physics loss as a secondary pre-training objective. MxGPS extends this direction by introducing multi-task specialisation as a structural inductive bias and by applying the BIM constraint across all task branches throughout training, not only during pre-training.

The central finding of this work is the identification of \emph{topology overfitting} as a structural failure mode in single-task GNN fine-tuning for power grids. We argue topology-agnostic behaviour is the more relevant criterion for a power grid foundation model, as operational value compounds when the model is deployed on grids absent from the training portfolio. The primary contributions of this work are as follows:
\begin{enumerate}
  \item We identify and characterise \textbf{topology overfitting}: the models with the lowest in-distribution PF error degrade the most under topology shift (GNS by $790\%$, GPS by over $1300\%$), an accuracy inversion exposed by our sliding-window cross-validation protocol (Section~\ref{sec:experiments}).
  \item We propose MxGPS, a multiplex graph transformer with $K$ task-specialised GPS branches over a shared encoder, jointly trained on SSE and PF with a shared BIM physics constraint, providing implicit regularisation against topology overfitting; a cross-branch attention module is evaluated in ablation (Section~\ref{sec:method}).
  \item We implement a two-phase training protocol: MGT self-supervised pre-training followed by multi-task fine-tuning with a linear probe phase (Section~\ref{sec:training}).
  \item We demonstrate that multi-task joint training addresses topology overfitting: MxGPS achieves $0\%$ BVR on all four zero-shot PF topologies and degrades by only $39\%$ in PF accuracy under topology shift, versus $190\%$--$1400\%$ for all single-task baselines (Sections~\ref{sec:experiments} and~\ref{sec:results}).
\end{enumerate}

\section{Problem Formulation}
\label{sec:problem}

\paragraph{Grid graph}
A power grid can be modelled as a graph $\mathcal{G} = (\mathcal{V}, \mathcal{E})$ where nodes represent buses and edges are lines and transformers connecting these buses. The graph is naturally heterogeneous since each node can have either no or multiple generators connected to it. Here, we transform the heterogeneous graph into a homogeneous one by representing node features with a fixed 14-dimensional feature vector that captures aggregate generator properties wherever they are present at a given bus $i \in \mathcal{V}$ in the grid:
\begin{equation}
\begin{split}
\mathbf{x}_i = [ & P_d,\; Q_d,\; P_g,\; Q_g,\; V_m,\; V_a,\; \mathbf{1}_\text{PQ}, \\
                 & \mathbf{1}_\text{PV},\; \mathbf{1}_\text{REF},\; V_m^{\min},\; V_m^{\max},\; G_s,\; B_s,\; V_n ]
\end{split}
\end{equation}
where $P_d, Q_d$ are active/reactive load; $P_g, Q_g$ active/reactive generation; $V_m, V_a$ voltage magnitude and angle; $\mathbf{1}_\text{PQ/PV/REF}$ are bus type one-hot flags; $V_m^{\min/\max}$ are per-bus operational voltage limits; $G_s, B_s$ are shunt admittance components; and $V_n$ is the nominal voltage. Transmission lines and transformers form edge set $\mathcal{E}$ with 11-dimensional edge features including branch power flows, admittance parameters $(Y_{ff}, Y_{ft}, Y_{tt}, Y_{tf})$, tap ratio, angle limits, and thermal rating.

\paragraph{AC power balance}
At every bus $i$, the AC power balance imposes the following constraints:
\begin{align}
P_i &= |V_i| \sum_{j \in \mathcal{N}(i)} |V_j|
        \bigl( G_{ij} \cos(\theta_i - \theta_j) + B_{ij} \sin(\theta_i - \theta_j) \bigr)
\label{eq:pbal} \\
Q_i &= |V_i| \sum_{j \in \mathcal{N}(i)} |V_j|
        \bigl( G_{ij} \sin(\theta_i - \theta_j) - B_{ij} \cos(\theta_i - \theta_j) \bigr)
\label{eq:qbal}
\end{align}
where $G_{ij} + jB_{ij}$ are entries of the bus admittance matrix and $\mathcal{N}(i)$ is the neighbourhood of bus $i$ in $\mathcal{G}$. Any physically valid operating point must satisfy Eqs.~\eqref{eq:pbal}--\eqref{eq:qbal}. Note: $V_a$ and $\theta_i$ denote the same voltage angle; $G_s, B_s$ in $\mathbf{x}_i$ are per-bus shunt admittances, while $G_{ij}, B_{ij}$ are mutual admittances.

\paragraph{Tasks}
We consider three tasks defined by different masking strategies applied to $\mathbf{x}_i$:
\begin{itemize}
  \item \textbf{Masked Grid Task~(MGT)}: self-supervised pre-training. Each feature dimension $f$ of each bus $i$ is masked independently with probability $\rho$; masked positions receive a learned $[\text{MASK}]$ token. The model is trained to reconstruct the original bus features from the unmasked context.
  \item \textbf{Power Flow~(PF)}: supervised regression. Given a known load profile and generator dispatch, predict voltage magnitudes and angles at all buses. Masking follows the physical observability structure: $V_m$ and $V_a$ are masked at PQ buses; $Q_g$ and $V_a$ at PV buses; $P_g$ and $Q_g$ at the reference (slack) bus, where the solver computes both to satisfy the power balance; $Q_g$ is similarly an output at PV buses, where reactive dispatch is determined by the voltage setpoint.
  \item \textbf{Static State Estimation~(SSE)}: supervised regression with noise. The ground truth is the clean AC power flow solution $\mathbf{x}^*$ produced by Newton-Raphson during dataset generation via \texttt{gridfm-datakit}~\cite{puech2025gridfm}. Each measurable bus feature ($P_d, Q_d, P_g, Q_g, V_m, V_a$) is first corrupted with additive Gaussian noise (std $\sigma_\varepsilon = 0.02$\,p.u.) and then independently dropped with probability $\rho_\text{mask} = 0.2$; dropped positions receive a learned \texttt{[MASK]} token. These corruption levels are deliberately chosen as a stress test of robustness to noisy and missing measurements, rather than as a calibrated model of a specific metering infrastructure. The model is trained to recover the full clean state $\mathbf{x}^*$ at \emph{all} buses, including both noisy and missing measurement positions.
\end{itemize}

\section{Methodology}
\label{sec:method}

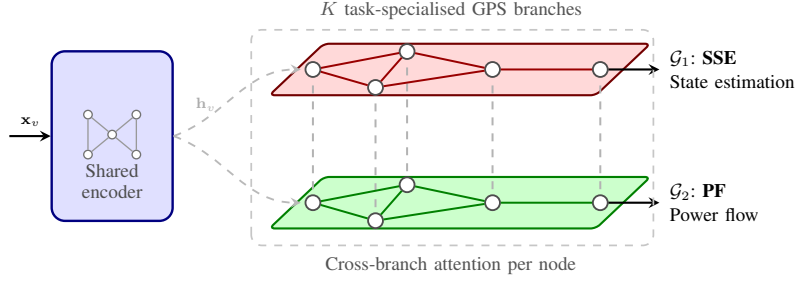
\begin{figure*}[t]
\centering
\begin{tikzpicture}[scale=0.8, transform shape,
    font=\small,
    bus/.style={circle, draw=black!70, fill=white, inner sep=0pt,
                minimum size=7pt, line width=0.7pt},
    tinybus/.style={circle, draw=black!55, fill=white, inner sep=0pt,
                    minimum size=4pt, line width=0.5pt},
    encbox/.style={rectangle, rounded corners=5pt,
                   draw=blue!50!black, fill=blue!12, line width=0.9pt,
                   minimum width=2.0cm, minimum height=2.8cm},
    arr/.style={->, >=stealth, line width=0.8pt},
    crossattn/.style={dashed, line width=0.75pt, color=gray!60},
    dasharr/.style={dashed, ->, >=stealth, line width=0.7pt, color=gray!55},
]

\filldraw[fill=red!15, draw=red!45!black, line width=0.8pt, rounded corners=2pt]
    (4.2,3.70) -- (9.6,3.70) -- (10.5,4.55) -- (5.1,4.55) -- cycle;

\filldraw[fill=green!20, draw=green!50!black, line width=0.8pt, rounded corners=2pt]
    (4.2,1.50) -- (9.6,1.50) -- (10.5,2.35) -- (5.1,2.35) -- cycle;

\draw[crossattn] (4.920,4.125) -- (4.920,1.925);
\draw[crossattn] (6.477,4.422) -- (6.477,2.222);
\draw[crossattn] (5.955,3.828) -- (5.955,1.628);
\draw[crossattn] (7.890,4.125) -- (7.890,1.925);
\draw[crossattn] (9.672,4.125) -- (9.672,1.925);

\node[bus] (l1n1) at (4.920,4.125) {};
\node[bus] (l1n2) at (6.477,4.422) {};
\node[bus] (l1n3) at (5.955,3.828) {};
\node[bus] (l1n4) at (7.890,4.125) {};
\node[bus] (l1n5) at (9.672,4.125) {};
\draw[red!60!black, line width=0.75pt]
    (l1n1)--(l1n2) (l1n1)--(l1n3) (l1n2)--(l1n3)
    (l1n2)--(l1n4) (l1n3)--(l1n4) (l1n4)--(l1n5);

\node[bus] (l2n1) at (4.920,1.925) {};
\node[bus] (l2n2) at (6.477,2.222) {};
\node[bus] (l2n3) at (5.955,1.628) {};
\node[bus] (l2n4) at (7.890,1.925) {};
\node[bus] (l2n5) at (9.672,1.925) {};
\draw[green!50!black, line width=0.75pt]
    (l2n1)--(l2n2) (l2n1)--(l2n3) (l2n2)--(l2n3)
    (l2n2)--(l2n4) (l2n3)--(l2n4) (l2n4)--(l2n5);

\node[encbox] (enc) at (1.6,3.025) {};

\node[tinybus] (en1) at (1.2,3.38) {};
\node[tinybus] (en2) at (2.0,3.38) {};
\node[tinybus] (en3) at (1.6,3.05) {};
\node[tinybus] (en4) at (1.2,2.72) {};
\node[tinybus] (en5) at (2.0,2.72) {};
\draw[black!40, line width=0.5pt]
    (en1)--(en3) (en2)--(en3) (en3)--(en4) (en3)--(en5) (en1)--(en4) (en2)--(en5);

\node[font=\small, align=center, text=black!70] at (1.6,2.28) {Shared\\encoder};

\draw[arr] (-0.1,3.025) -- (enc.west);
\node[font=\scriptsize, above=1pt] at (0.25,3.025) {$\mathbf{x}_v$};

\draw[dasharr] (enc.east) to[out=20,in=180] ($(l1n1)+(-0.2,0)$);
\draw[dasharr] (enc.east) to[out=-20,in=180] ($(l2n1)+(-0.2,0)$);
\node[font=\scriptsize, text=gray!65] at (3.15,3.55) {$\mathbf{h}_v$};

\draw[arr] (l1n5) -- (10.65,4.125);
\node[font=\small, align=left, anchor=west] at (10.70,4.125)
    {$\mathcal{G}_1$: \textbf{SSE}\\[1pt]State estimation};

\draw[arr] (l2n5) -- (10.65,1.925);
\node[font=\small, align=left, anchor=west] at (10.70,1.925)
    {$\mathcal{G}_2$: \textbf{PF}\\[1pt]Power flow};

\draw[gray!45, dashed, rounded corners=4pt, line width=0.65pt]
    (3.9,1.22) rectangle (10.55,4.78);

\node[font=\small, text=gray!65!black] at (7.2,5.10)
    {$K$ task-specialised GPS branches};

\node[font=\small, text=gray!65!black] at (7.2,0.88)
    {Cross-branch attention per node};

\end{tikzpicture}
\caption{MxGPS architecture with $K{=}2$ branches. A shared node encoder maps raw bus features $\mathbf{x}_v$ and structural positional encodings (RWSE, LapPE) to a common embedding $\mathbf{h}_v$, which is broadcast to both task-specialised GPS branches. Each branch processes the identical physical graph topology through its own GPS layers, conditioned by a learnable task token $\mathbf{t}_k$. Dashed vertical lines indicate the optional cross-branch attention (MxGPS-cross only), which exchanges per-node representations between branches at each GPS layer, providing mutual regularisation across tasks.}
\label{fig:architecture}
\end{figure*}

MxGPS has three components: a shared node encoder, $K$ task-specialised GPS branches, and an optional cross-branch attention module evaluated in ablation (absent in MxGPS-ind, present in MxGPS-cross). In this work $K=2$ (SSE and PF branches); the design is intended to extend to larger task portfolios by adding branches with new masking transforms and prediction heads, though we do not evaluate $K>2$ here and treat it as future work. Figure~\ref{fig:architecture} illustrates the architecture.

\subsection{Shared Node Encoder}
\label{sec:encoder}

A single linear projection maps raw bus features and structural positional encodings to a shared $d$-dimensional embedding, providing a common physical representation before task-specific processing:
\begin{equation}
\mathbf{h}_i^{(0)} = \mathbf{W}_\text{in}\, \mathbf{x}_i
                   + \mathbf{W}_\text{pe}
                     \begin{bmatrix} \text{RWSE}_i \\ \text{LapPE}_i \end{bmatrix}
\label{eq:shared_enc}
\end{equation}
where $\text{RWSE}_i \in \mathbb{R}^{r}$ is the $r$-step Random Walk Structural Encoding~\cite{dwivedi2022graph} and $\text{LapPE}_i \in \mathbb{R}^{2p}$ is the top-$p$ eigenvectors of the normalised graph Laplacian. Both encodings capture local and global structural roles of buses independent of their feature values, enabling cross-topology transfer~\cite{hamann2024foundation,rampasek2022recipe}. Both $\mathbf{W}_\text{in} \in \mathbb{R}^{d \times 14}$ and $\mathbf{W}_\text{pe} \in \mathbb{R}^{d \times (r+2p)}$ are learnable weight matrices, where $d$ is the hidden embedding dimension (distinct from the demand subscript in $P_d, Q_d$). Masked input positions receive a learned $[\text{MASK}]$ token embedding $\mathbf{m}$ in place of $\mathbf{W}_\text{in}\,\mathbf{x}_i$, following Devlin et al.~\cite{devlin2019bert} and GridFM.

\subsection{Task Token Conditioning}
\label{sec:tokens}

A learnable task token $\mathbf{t}_k \in \mathbb{R}^d$ is added to the shared encoder output before branch $k$ processes its input:
\begin{equation}
\mathbf{h}_i^{(0,k)} = \mathbf{h}_i^{(0)} + \mathbf{t}_k
\label{eq:task_token}
\end{equation}
This conditions each branch on its task identity without any architectural modification, analogous to task-prefix conditioning in language models; specialisation emerges from task-specific gradient steering through the branch-specific loss.

\subsection{Task-Specialised GPS Branches}
\label{sec:branches}

Each branch $k \in \{1, \ldots, K\}$ applies $L$ GPS layers~\cite{rampasek2022recipe} to $\mathbf{h}^{(0,k)}$. Each GPS layer combines a local GatedGCN message-passing step~\cite{bresson2017residual} with a global Transformer self-attention step~\cite{vaswani2017attention}.

\textbf{GatedGCN (local path).}
Edge gates $\hat{\eta}_{ij}$ are computed from node and edge embeddings and normalised over each node's neighbourhood, where BN denotes Batch Normalisation and $\sigma$ is the sigmoid function:
\begin{align}
\mathbf{e}_{ij}^{(l+1)} &= \mathbf{e}_{ij}^{(l)} + \text{ReLU}\!\left(\text{BN}\!\left(
  \mathbf{W}_1 \mathbf{h}_i^{(l)} + \mathbf{W}_2 \mathbf{h}_j^{(l)} +
  \mathbf{W}_3 \mathbf{e}_{ij}^{(l)}\right)\right) \label{eq:edge_update}\\
\hat{\eta}_{ij} &= \frac{\sigma(\mathbf{e}_{ij}^{(l+1)})}
  {\sum_{j' \in \mathcal{N}(i)} \sigma(\mathbf{e}_{ij'}^{(l+1)}) + \varepsilon}
\label{eq:gate_norm}\\
\mathbf{h}_i^{(l+\frac{1}{2})} &= \mathbf{h}_i^{(l)} + \text{ReLU}\!\left(\text{BN}\!\left(
  \sum_{j \in \mathcal{N}(i)} \hat{\eta}_{ij} \odot \mathbf{W}_h \mathbf{h}_j^{(l)}\right)\right)
\label{eq:node_update}
\end{align}

\textbf{Global Transformer with edge bias (global path).}
Full $O(N^2)$ self-attention is applied over all buses in the graph; the attention logit between $i$ and $j$ is additively shifted by a linear projection of the edge embedding, allowing the Transformer to be directly aware of the physical connection between each bus pair:
\begin{equation}
a_{ij}^{(l)} = \frac{\bigl(\mathbf{Q}\,\mathbf{h}_i^{(l+\frac{1}{2})}\bigr)^\top
  \bigl(\mathbf{K}\,\mathbf{h}_j^{(l+\frac{1}{2})}\bigr)}{\sqrt{d_h}}
  + \phi\!\left(\mathbf{e}_{ij}^{(l+1)}\right)
\label{eq:edge_bias_attn}
\end{equation}
where $\phi : \mathbb{R}^{d_e} \to \mathbb{R}^{n_\text{heads}}$ produces per-head additive logit biases, following the GraphGPS formulation~\cite{rampasek2022recipe}.

\textbf{GPS layer output.}
Local and global paths are fused through residual connections and LayerNorm, followed by a position-wise Feed-Forward Network~(FFN) with Gaussian Error Linear Unit~(GELU) activation, where MHA denotes multi-head attention:
\begin{equation}
\mathbf{h}_i^{(l+1)} = \text{LN}\!\left(\mathbf{h}_i^{(l+\frac{1}{2})}
  + \text{MHA}\!\left(\mathbf{h}^{(l+\frac{1}{2})}, a^{(l)}\right)
  + \text{FFN}\!\left(\cdot\right)\right)
\end{equation}

\subsection{Cross-Branch Attention}
\label{sec:crossattn}

When cross-branch attention is enabled, after each GPS layer $l$ the representations of all $K$ branches are exchanged per node:
\begin{equation}
\tilde{\mathbf{h}}_i^{(k,l)} = \mathbf{h}_i^{(k,l)} + \text{CrossAttn}\!\left(
  \mathbf{h}_i^{(k,l)},\; \bigl[\mathbf{h}_i^{(k',l)}\bigr]_{k' \neq k}\right)
\label{eq:crossattn}
\end{equation}
where CrossAttn uses branch $k$'s representation as query and all other branches as keys and values.
This is applied independently per node with complexity $O(K^2 \cdot d)$ per layer, lightweight for $K=2$ and tractable as the task portfolio grows, allowing task experts to leverage complementary representations and propagate cross-task information through subsequent message-passing rounds.

\subsection{Task Prediction Heads}
\label{sec:heads}

Each branch $k$ has its own two-layer MLP regression head producing per-bus predictions $\hat{\mathbf{y}}_i^{(k)} \in \mathbb{R}^4$ covering active and reactive power generation ($\hat{P}_{g,i}, \hat{Q}_{g,i}$) and voltage state ($\hat{V}_{m,i}, \hat{V}_{a,i}$).

\subsection{Loss Function}
\label{sec:loss}

The total training loss combines per-task supervised losses with an optional physics-informed self-supervised term and a boundary penalty:
\begin{equation}
\mathcal{L} = \sum_{k=1}^{K} \lambda_k \mathcal{L}_k^{\text{task}}
            + \lambda_{\text{phys}} \mathcal{L}_{\text{phys}}
            + \lambda_{\text{bnd}} \mathcal{L}_{\text{bnd}}
\label{eq:loss}
\end{equation}

\textbf{MGT pre-training loss.}
For the self-supervised pre-training task we use the Scaled Cosine Error~(SCE) from GraphMAE~\cite{hou2022graphmae} over masked bus feature positions $\mathcal{M}$:
\begin{equation}
\mathcal{L}_{\text{SCE}} = \frac{1}{|\mathcal{M}|} \sum_{(i,f) \in \mathcal{M}}
  \left(1 - \frac{\hat{x}_{if}\, x_{if}}{\|\hat{\mathbf{x}}_i\|\, \|\mathbf{x}_i\|}\right)^\gamma
\end{equation}
where $\gamma = 2.0$ is a sharpening exponent that up-weights hard examples. SCE is preferred over MSE because it penalises directional rather than absolute deviation, better suiting normalised node features~\cite{hou2022graphmae}.

\textbf{Supervised task loss.}
For PF and SSE, a dimension-weighted mean squared error is applied over the task-specific masked prediction positions, matching the gridfm-graphkit evaluation protocol:
\begin{equation}
\mathcal{L}_k^{\text{task}} = \sum_{f \in \{P_g, Q_g, V_m, V_a\}} w_f
  \cdot \frac{1}{|\mathcal{M}_k|} \sum_{i \in \mathcal{M}_k} \bigl(\hat{y}_{if} - y_{if}\bigr)^2
\end{equation}

\textbf{Physics SSL loss~(BIM).}
The Bus Injection Model~(BIM) residual is applied to the predicted voltages of every branch simultaneously, providing a shared physics constraint:
\begin{equation}
\mathcal{L}_{\text{phys}} = \frac{1}{K} \sum_{k=1}^{K} \frac{1}{N} \sum_{i=1}^{N}
  \Bigl[\bigl(\hat{P}_i^{(k)} - P_i^{\text{BIM}}\bigr)^2 +
        \bigl(\hat{Q}_i^{(k)} - Q_i^{\text{BIM}}\bigr)^2\Bigr]
\end{equation}
where $P_i^{\text{BIM}}, Q_i^{\text{BIM}}$ are computed from predicted voltages via Eqs.~\eqref{eq:pbal}--\eqref{eq:qbal} using the per-case bus admittance matrix. This is a key advantage over GridFM~\cite{hamann2024foundation}: the physics constraint is applied to all task experts simultaneously, not only during the reconstruction pre-training phase.

\textbf{Boundary loss.}
A soft penalty discourages voltage magnitudes outside operational limits $[V_{m,i}^{\min}, V_{m,i}^{\max}]$ available in the bus feature vector:
\begin{equation}
\begin{split}
\mathcal{L}_{\text{bnd}} &= \frac{1}{K} \sum_{k=1}^{K} \frac{1}{N} \sum_{i=1}^{N}
  \Bigl[\text{ReLU}\!\bigl(V_{m,i}^{\min} - \hat{V}_{m,i}^{(k)}\bigr)^2 \\
  &\quad + \text{ReLU}\!\bigl(\hat{V}_{m,i}^{(k)} - V_{m,i}^{\max}\bigr)^2\Bigr]
\end{split}
\end{equation}

\section{Training Protocol}
\label{sec:training}

We follow a two-phase training protocol analogous to GridFM's linear probe fine-tuning curriculum~\cite{hamann2024foundation}.

\textbf{Phase~A: MGT pre-training.}
A single GPS model is pre-trained on the Masked Grid Task using the SCE loss plus a layered physics loss applied at each GPS layer, producing warm, general-purpose encoder and GPS layer weights before any task-specific gradient signal is introduced.

\textbf{Phase~B - MxGPS: multi-task fine-tuning.}
The pre-trained encoder and GPS layer weights are loaded into all $K$ branches of MxGPS (all branches are initialised identically from the Phase~A checkpoint). The shared encoder is frozen for the first 30 epochs while only task tokens and prediction heads are trained (linear probe phase); the encoder is then unfrozen and the early-stopping budget resets for full fine-tuning, mirroring GridFM's linear probe protocol.

\textbf{Phase~B - Single-task baselines: task-specific fine-tuning.}
The same two-phase protocol is applied to all five single-task GNN baselines (GCN, GAT, GNS, GPS, and GNS-kit; see Section~\ref{sec:baselines}) on PF and SSE. An MGT-pre-trained checkpoint is loaded via a weight transfer step that copies all tensors whose name and shape match the target model; the transfer is complete for all architectures evaluated here (GNS-kit's physics decoder heads are parameter-free). Single-task Phase~B is trained end-to-end from the warm-started weights with no freeze phase, since all parameters are pre-trained.

\section{Experiments}
\label{sec:experiments}

\subsection{Datasets}
\label{sec:data}

We generate operational scenarios from standard MATPOWER test systems~\cite{zimmerman2010matpower} (including grids from the PGLib-OPF benchmark~\cite{babaeinejadsarookolaee2019power}) using \texttt{gridfm-datakit}~\cite{puech2025gridfm}, which applies a hybrid load perturbation strategy (global temporal scaling from real load profiles combined with per-bus noise) followed by Newton-Raphson AC power flow~\cite{stott1974review}. Data is stored in Parquet format with per-bus, per-generator, and per-branch state variables. Approximately $10\,000$ scenarios are generated per case, and an 80\,/\,10\,/\,10\% train-validation-test split is applied deterministically by scenario ID.

\textbf{Evaluation protocol.}
We order the eight IEEE cases~\cite{zimmerman2010matpower} by size (14, 24, 30, 57, 73, 118, 162, 300~buses) and form three overlapping training windows of six consecutive cases: fold~1 trains on the 14--118-bus cases and holds out \{162, 300\}; fold~2 trains on 24--162 and holds out \{14, 300\}; fold~3 trains on 30--300 and holds out \{14, 24\}. Across folds this yields four unique zero-shot topologies: the 14- and 300-bus cases are zero-shot in two folds each, the 24- and 162-bus cases in one. MGT pre-training (Phase~A) is repeated per fold on that fold's six training cases, so no zero-shot topology is seen at any training stage of the corresponding fold. For zero-shot cases, the per-case normaliser is fitted on a single representative scenario from that case rather than a training split, since zero-shot topologies contribute no training data and their full scenario set is reserved for evaluation; as with training cases, power features are scaled by the fixed per-case base MVA and only the nominal-voltage constant is inferred from the fitted scenario, following the per-case normaliser convention of \texttt{gridfm-datakit}~\cite{puech2025gridfm}. Reported accuracy, BVR, and degradation figures are averaged over the folds in which each topology is zero-shot; the full scenario set of each zero-shot case is used for evaluation.

\textbf{Normalisation.}
Bus power features are normalised by the per-case base apparent power (100~MVA for all cases). Voltage angles are converted from degrees to radians. A per-case normaliser is fitted on the training split and applied consistently to all evaluation sets~\cite{puech2025gridfm}.

\subsection{Models and Baselines}
\label{sec:baselines}

We compare MxGPS against five single-task GNN baselines (each trained independently per task) and one analytical physics baseline. GNN baselines follow the two-phase GridFM protocol: MGT pre-training (Phase~A) followed by task-specific fine-tuning (Phase~B).

\textbf{Physics baseline (DC~PF).}
The linearised DC power-flow approximation: given bus active injections ($P_g - P_d$), it solves $\mathbf{B}_\text{red}\,\boldsymbol{\theta} = \mathbf{P}_\text{inj}$ via sparse linear algebra for the voltage angles, while setting all voltage magnitudes to $1.0$\,p.u. The $\mathbf{B}$ matrix is reconstructed from the dataset's edge admittance features. Applied to both PF and SSE in-distribution settings. Newton-Raphson AC power flow was also implemented but excluded due to convergence failures on a subset of test cases.

\begin{table*}[!t]
\centering
\caption{MGT in-distribution accuracy and calibration. MAE/RMSE in p.u.\ except $\text{MAE}_{V_a}$ (rad); Slope$=1$, $R^2=1$ ideal. Best per column in \textbf{bold}.}
\label{tab:mgt_acc}
\footnotesize
\begin{tabular}{lcccccccc}
\toprule
Model & $\text{MAE}_{V_m}$ & $\text{MAE}_{V_a}$ & $\text{RMSE}_{V_m}$ & BVR\,(\%) & Slope$_{V_m}$ & $R^2_{V_m}$ & Slope$_{V_a}$ & $R^2_{V_a}$ \\
\midrule
Naive ($V_m{=}1$) & 0.0333 & --- & 0.0369 & 0.00 & 0.000 & 0.000 & --- & --- \\
\midrule
GCN     & 0.0230 & \textbf{0.1213} & \textbf{0.0267} & \textbf{0.13} & 0.202 & \textbf{0.067} & 0.127 & \textbf{0.149} \\
GAT     & 0.0358 & 0.1922 & 0.0448 & 26.97 & \textbf{0.378} & 0.044 & 0.089 & 0.009 \\
GPS     & \textbf{0.0228} & 0.2010 & 0.0282 & 14.93 & 0.019 & 0.000 & $-$0.002 & 0.000 \\
GNS     & 0.0445 & 0.1304 & 0.0586 & 19.76 & $-$0.335 & 0.025 & \textbf{0.252} & 0.123 \\
GNS-kit & 0.0315 & 0.1398 & 0.0427 & 3.47 & 0.000 & 0.000 & 0.146 & 0.096 \\
\bottomrule
\end{tabular}
\end{table*}

\textbf{GNN baselines.}

\begin{itemize}
  \item \textbf{GCN}~\cite{kipf2017semi}: Homogeneous graph convolutional network, 4~layers, hidden dimension~128.
  \item \textbf{GAT}~\cite{velickovic2018graph}: Heterogeneous graph attention network with GATv2 convolutions~\cite{brody2022attentive}, 4~layers, 4~heads, hidden dimension~128.
  \item \textbf{GNS}: Heterogeneous graph network with TransformerConv message passing over bus, generator, and edge node types; 6~layers, 8~heads, hidden dimension~256.
  \item \textbf{GPS}~\cite{rampasek2022recipe}: Single-task GPS (GatedGCN + global Transformer with edge-biased attention); 4~layers, 4~heads, hidden dimension~128, RWSE ($r=16$) + LapPE ($p=8$). GPS shares the same backbone architecture and hyperparameters as each individual MxGPS branch, making the GPS vs.\ MxGPS comparison a controlled ablation of single-task vs.\ joint multi-task training.
  \item \textbf{GNS-kit}: Exact replica of the gridfm-graphkit HGNS architecture~\cite{hamann2024foundation,puech2025gridfm}; 12~layers, 8~heads, hidden dimension~48, trained with SCE + layered physics loss and gridfm-graphkit hyperparameters. The name distinguishes this model from the GridFM project as a whole; it is the closest proxy to GridFM we can build without access to the original pre-training corpus and checkpoints.
\end{itemize}

MxGPS is evaluated in two configurations: (i) \textbf{MxGPS-ind}, with $K=2$ independent branches (SSE and PF) and no cross-branch attention, which we adopt as the \emph{primary} configuration and refer to simply as MxGPS where unambiguous; and (ii) \textbf{MxGPS-cross}, which adds cross-branch attention after each GPS layer and serves as an ablation of that component. Both variants use hidden dimension~128, 4~GPS layers, 4~attention heads, and RWSE ($r=16$) + LapPE ($p=8$) structural encodings, trained jointly on SSE and PF batches in round-robin order.

\subsection{Metrics}
\label{sec:metrics}

We report the following metrics per task:
(1) $\text{MAE}_{V_m}$: mean absolute error on voltage magnitude (p.u.);
(2) $\text{RMSE}_{V_m}$: root mean squared error on voltage magnitude;
(3) $\text{MAE}_{V_a}$: mean absolute error on voltage angle (rad);
(4) \textbf{BVR\,(\%)}: Boundary Violation Rate, defined as the fraction of bus predictions where $\hat{V}_{m,i} \notin [V_{m,i}^{\min}, V_{m,i}^{\max}]$, measuring operational safety;
(5) \textbf{Regression slope} and $R^2$: following Yang et al.~\cite{yang2026gridsfm}, we fit a linear regression of predicted on true values per output channel and report the slope (ideally $1.0$, indicating correct magnitude) and the Pearson $R^2$ (fraction of variance explained), detecting models that collapse toward the mean despite low MAE.

\subsection{Implementation Details}
\label{sec:impl}

All models are implemented in PyTorch with PyTorch Geometric~\cite{fey2019fast}. Experiments are tracked with MLflow. Training uses the AdamW optimiser~\cite{loshchilov2019decoupled} with $\beta_1 = 0.9$, $\beta_2 = 0.999$, and weight decay $10^{-4}$. Learning rate is reduced on validation loss plateau (factor 0.5, patience 10~epochs). Maximum training is 500~epochs with early stopping (patience 50) for all models. GNS-kit uses graphkit-matched hyperparameters: lr $= 5 \times 10^{-4}$, decay factor~0.7, patience~5. All models are trained on a single GPU (L40S) with batch size~64 (32 for MxGPS). Seed is fixed for reproducibility. Single-task baselines are fine-tuned with only the supervised task loss: SSE uses dimension-weighted MAE with $w_f = [0.10, 0.10, 0.35, 0.35]$ for $[P_g, Q_g, V_m, V_a]$, and PF uses $0.90 \cdot \text{MSE} + 0.10 \cdot \mathcal{L}_{\text{phys}}$, where the physics term is a per-layer residual exposed only by GNS-kit's architecture and is therefore zero for GCN, GAT, GPS, and GNS; no boundary penalty $\mathcal{L}_{\text{bnd}}$ is applied to any single-task baseline. MxGPS is the only model trained with the full objective of Eq.~\eqref{eq:loss}, using per-task weights $\lambda_k = 1.0$ (SSE and PF), $\lambda_{\text{phys}} = 0.01$, and $\lambda_{\text{bnd}} = 0.01$, with the same dimension weights $w_f$ above. Base learning rate is $1 \times 10^{-3}$ for MxGPS-ind and $5 \times 10^{-4}$ for MxGPS-cross. MGT pre-training (Phase~A) masks each of the first six bus features independently with probability $\rho = 0.5$, matching \texttt{gridfm-graphkit}'s random masking scheme for this stage.

\begin{figure}[h]
\centering
\includegraphics[width=\columnwidth]{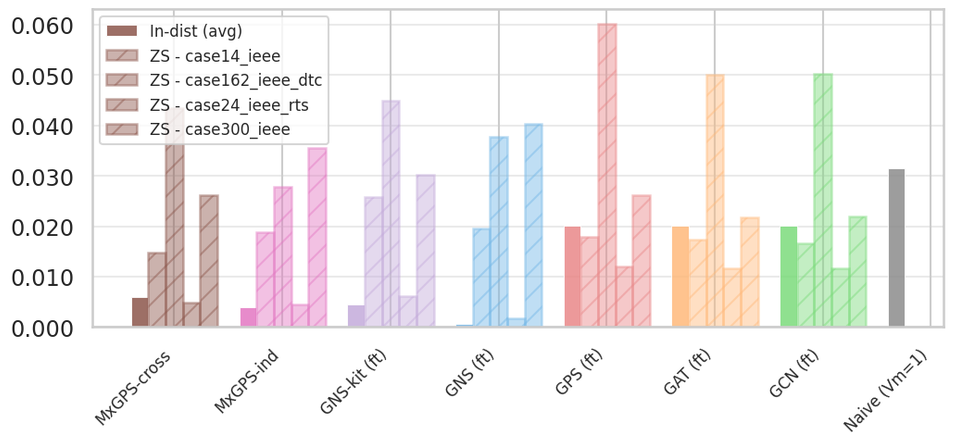}
\caption{SSE zero-shot MAE$_{V_m}$ across the four unseen topologies. Among models with meaningful in-distribution SSE accuracy (GNS, GNS-kit, MxGPS), MxGPS variants degrade the least (Table~\ref{tab:degradation}); GPS, GAT, and GCN show artificially low degradation due to in-distribution magnitude collapse (Table~\ref{tab:sse_acc}).}
\label{fig:zeroshot}
\end{figure}

\section{Results and Discussion}
\label{sec:results}

\subsection{Masked Grid Task Pre-training}
\label{sec:mgt_results}

Table~\ref{tab:mgt_acc} serves one purpose: to show that MGT pre-training carries essentially no per-scenario voltage magnitude signal, motivating the supervised Phase~B fine-tuning. It reports in-distribution MGT accuracy and calibration after Phase~A, averaged over the held-out test splits of each fold's training cases; the Naive ($V_m{=}1$) row is a lower-bound reference. GPS, GCN, and GNS-kit edge out the Naive predictor on $\text{MAE}_{V_m}$, though GNS-kit's margin is marginal ($0.0315$ vs.\ $0.0333$); calibration is poor, confirming that the MGT objective does not directly optimise for voltage recovery. Rather than relying solely on this self-supervised signal, MxGPS introduces supervised SSE and PF gradients in Phase~B, directly training each branch on the quantities it predicts at inference.

\subsection{Power Flow}
\label{sec:pf_results}

Table~\ref{tab:pf_acc} reports in-distribution PF accuracy and calibration. All single-task baselines use the two-phase protocol (Phase~A MGT pre-training $\to$ Phase~B task-specific fine-tuning); MxGPS variants are jointly fine-tuned on SSE and PF simultaneously.

GNS and GPS achieve the lowest in-distribution $\text{MAE}_{V_m}$ ($0.0022$ and $0.0023$, respectively), but at the cost of substantial boundary violations (BVR $9.4\%$ and $7.3\%$). MxGPS variants show higher in-distribution error ($\text{MAE}_{V_m} \approx 0.021$) but near-zero BVR ($\leq 0.33\%$), reflecting the regularising effect of joint multi-task training on operational constraint satisfaction. On calibration, single-task baselines achieve near-perfect regression slope ($>0.94$), while MxGPS exhibits partial $V_m$ magnitude suppression (slope $\approx 0.5$--$0.6$), an area for future improvement. Because GPS shares the same backbone architecture and hyperparameters as each individual MxGPS branch, the GPS vs.\ MxGPS comparison on this table isolates the effect of multi-task joint training from any architectural difference.

\begin{figure*}[h]
\centering
\includegraphics[width=0.86\textwidth]{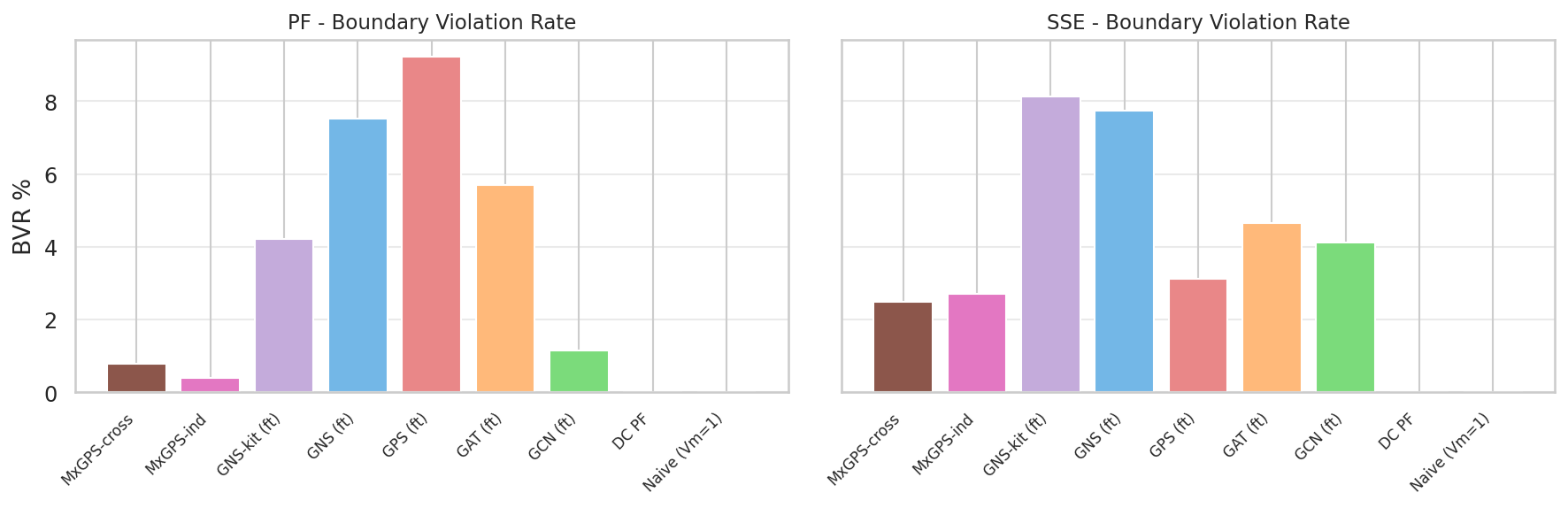}
\caption{In-distribution boundary violation rate per model and task. MxGPS variants consistently achieve the lowest violation rates, particularly on PF.}
\label{fig:multitask}
\end{figure*}

\begin{table*}[h]
\centering
\caption{PF in-distribution accuracy and calibration (mean over 3 cross-validation folds). MAE/RMSE in p.u.\ except $\text{MAE}_{V_a}$ (rad). Calibration: regression slope and $R^2$ (Slope$=1$, $R^2=1$ ideal; best Slope is closest to~1.0). MxGPS variants are jointly fine-tuned on SSE$+$PF; single-task models independently on PF. Best per column among GNN and MxGPS models in \textbf{bold}. DC~PF separated by rule: sets $V_m{=}1.0$\,p.u.\ throughout, so Slope$_{V_m}$/$R^2_{V_m}=0$; Slope$_{V_a}$/$R^2_{V_a}$ not computed.}
\label{tab:pf_acc}
\footnotesize
\begin{tabular}{lcccccccc}
\toprule
Model & $\text{MAE}_{V_m}$ & $\text{MAE}_{V_a}$ & $\text{RMSE}_{V_m}$ & BVR\,(\%) & Slope$_{V_m}$ & $R^2_{V_m}$ & Slope$_{V_a}$ & $R^2_{V_a}$ \\
\midrule
DC PF & 0.0286 & 0.2201 & 0.0322 & 0.00 & 0.000 & 0.000 & --- & --- \\
\midrule
GCN     & 0.0043 & 0.0175 & 0.0055 & 1.22  & 0.942 & 0.974 & 0.982 & 0.990 \\
GAT     & 0.0027 & 0.0161 & 0.0036 & 5.48  & 0.980 & 0.987 & 0.990 & 0.992 \\
GPS     & 0.0023 & \textbf{0.0057} & \textbf{0.0030} & 7.29  & 1.009 & 0.989 & 1.008 & \textbf{0.999} \\
GNS     & \textbf{0.0022} & 0.0184 & 0.0031 & 9.41  & \textbf{0.996} & \textbf{0.995} & \textbf{0.995} & 0.995 \\
GNS-kit & 0.0064 & 0.0173 & 0.0115 & 4.41  & 1.026 & 0.770 & 0.978 & 0.992 \\
\midrule
MxGPS-ind   & 0.0218 & 0.0464 & 0.0243 & \textbf{0.23} & 0.480 & 0.400 & 0.797 & 0.958 \\
MxGPS-cross & 0.0207 & 0.0530 & 0.0227 & 0.33  & 0.578 & 0.558 & 0.808 & 0.967 \\
\bottomrule
\end{tabular}
\end{table*}

\subsection{Static State Estimation}
\label{sec:sse_results}

Table~\ref{tab:sse_acc} reports in-distribution SSE accuracy and calibration under the same two-phase protocol. The calibration columns reveal a critical failure mode: GPS, GAT, and GCN show near-zero $V_m$ regression slope ($< 0.1$) and $R^2 < 0.1$, confirming they collapse to predicting the mean voltage despite $\text{MAE}_{V_m} \approx 0.020$ that may appear competitive. This is a low-MAE, low-information failure mode that bare accuracy statistics do not expose, directly motivating the calibration protocol of Yang et al.~\cite{yang2026gridsfm}: a model predicting the mean achieves acceptable average error while providing zero per-bus voltage information. MxGPS variants do not exhibit this collapse (slope $\approx 0.92$, $R^2 \approx 0.98$), consistent with the joint SSE$+$PF gradient signal preventing mode collapse to the mean. GNS achieves both the lowest $\text{MAE}_{V_m}$ ($0.0007$) and near-perfect calibration ($\text{slope} = 0.999$, $R^2 = 0.999$), reflecting its heterogeneous GNN design.

\begin{table*}[h]
\centering
\caption{SSE in-distribution accuracy and calibration (mean over 3 cross-validation folds). MAE/RMSE in p.u.\ except $\text{MAE}_{V_a}$ (rad). Calibration: regression slope and $R^2$ (Slope$=1$, $R^2=1$ ideal; best Slope is closest to~1.0). $\dagger$: GPS, GAT, GCN collapse on $V_m$ (slope and $R^2 < 0.1$), indicating mean prediction rather than genuine voltage estimation. GPS also collapses on $V_a$. Best per column among GNN and MxGPS models in \textbf{bold}. DC~PF separated by rule: sets $V_m{=}1.0$\,p.u.\ throughout; Slope$_{V_a}$/$R^2_{V_a}$ not computed.}
\label{tab:sse_acc}
\footnotesize
\begin{tabular}{lcccccccc}
\toprule
Model & $\text{MAE}_{V_m}$ & $\text{MAE}_{V_a}$ & $\text{RMSE}_{V_m}$ & BVR\,(\%) & Slope$_{V_m}$ & $R^2_{V_m}$ & Slope$_{V_a}$ & $R^2_{V_a}$ \\
\midrule
DC PF & 0.0315 & 0.8118 & 0.0354 & 0.00 & 0.000 & 0.000 & --- & --- \\
\midrule
GCN$^\dagger$  & 0.0201 & 0.0076 & 0.0240 & 2.88 & 0.076 & 0.083 & 0.990 & 0.998 \\
GAT$^\dagger$  & 0.0202 & 0.0079 & 0.0240 & 2.86 & 0.079 & 0.084 & 0.965 & 0.998 \\
GPS$^\dagger$  & 0.0201 & 0.0918 & 0.0241 & 2.90 & 0.091 & 0.101 & 0.177 & 0.194 \\
\midrule
GNS          & \textbf{0.0007} & \textbf{0.0030} & \textbf{0.0011} & 6.33          & \textbf{0.999} & \textbf{0.999} & \textbf{1.000} & \textbf{1.000} \\
GNS-kit      & 0.0045          & 0.0132          & 0.0104          & 7.67          & 0.953          & 0.838          & 1.003          & 0.968 \\
MxGPS-ind    & 0.0039          & 0.0258          & 0.0058          & \textbf{1.53} & 0.926          & 0.983          & 0.900          & 0.988 \\
MxGPS-cross  & 0.0060          & 0.0228          & 0.0081          & 1.84          & 0.920          & 0.983          & 0.961          & 0.991 \\
\bottomrule
\end{tabular}
\end{table*}

\subsection{Zero-Shot Transfer and Topology Overfitting}
\label{sec:multitask_results}

Figure~\ref{fig:multitask} shows the in-distribution boundary violation rate across models and tasks; MxGPS variants achieve the lowest BVR on PF ($\leq 0.33\%$). Table~\ref{tab:zeroshot_pf} reports zero-shot results across all four unseen topologies (14-, 24-, 162-, and 300-bus) for both PF and SSE, averaged over the applicable cross-validation folds, and Figure~\ref{fig:zeroshot} shows zero-shot SSE $\text{MAE}_{V_m}$ per model across all four topologies.

The strongest evidence sits on the two larger zero-shot systems, which are the closest analogue of foundation-model deployment on unseen grids. On the 162-bus case, MxGPS attains both the lowest zero-shot PF $\text{MAE}_{V_m}$ overall ($0.0230$) and $0\%$ BVR; on the 300-bus case, its error ($0.0247$) is within $0.006$~p.u.\ of the best baseline (GNS-kit, $0.0185$), whose predictions however violate voltage limits at $4.7\%$ of buses. On the two small zero-shot systems (14- and 24-bus), several baselines retain lower absolute MAE, so MxGPS's advantage is concentrated where grids are larger and structurally richer. Crucially, MxGPS is the only model family with $\text{BVR}=0\%$ on \emph{all four} zero-shot PF topologies, while GPS incurs up to $37\%$ BVR on case300 and GNS up to $28\%$ on case24.

\begin{table*}[h]
\centering
\caption{Zero-shot transfer to four unseen topologies for PF and SSE. $\text{MAE}_{V_m}$ in p.u., averaged over the folds in which each topology is zero-shot. Best $\text{MAE}_{V_m}$ per topology and task in \textbf{bold}; $0\%$ BVR entries also \textbf{bold}.}
\label{tab:zeroshot_pf}
\small
\resizebox{0.8\textwidth}{!}{%
\begin{tabular}{lcccccccc}
\toprule
 & \multicolumn{2}{c}{14-bus} & \multicolumn{2}{c}{24-bus} & \multicolumn{2}{c}{162-bus} & \multicolumn{2}{c}{300-bus} \\
\cmidrule(lr){2-3}\cmidrule(lr){4-5}\cmidrule(lr){6-7}\cmidrule(lr){8-9}
Model & $\text{MAE}_{V_m}$ & BVR\,(\%) & $\text{MAE}_{V_m}$ & BVR\,(\%) & $\text{MAE}_{V_m}$ & BVR\,(\%) & $\text{MAE}_{V_m}$ & BVR\,(\%) \\
\midrule
\multicolumn{9}{l}{\textit{Power Flow}} \\
GCN         & \textbf{0.0128} & \textbf{0.00} & 0.0053 & 0.03  & 0.0304 & 1.13  & 0.0296 & 2.26 \\
GAT         & 0.0164 & 12.64 & 0.0031 & 3.12  & 0.0358 & 5.81  & 0.0295 & 3.87 \\
GPS         & 0.0266 & \textbf{0.00} & 0.0037 & 4.10  & 0.0646 & 14.66 & 0.0411 & 37.04 \\
GNS         & 0.0158 & 10.91 & \textbf{0.0025} & 28.17 & 0.0312 & 0.39  & 0.0305 & 6.17 \\
GNS-kit     & 0.0195 & 0.30  & 0.0102 & 8.80  & 0.0264 & 3.06  & \textbf{0.0185} & 4.74 \\
MxGPS-ind   & 0.0383 & \textbf{0.00} & 0.0354 & \textbf{0.00} & \textbf{0.0230} & \textbf{0.00} & 0.0247 & \textbf{0.00} \\
MxGPS-cross & 0.0350 & \textbf{0.00} & 0.0340 & \textbf{0.00} & 0.0271 & \textbf{0.00} & 0.0342 & \textbf{0.00} \\
\midrule
\multicolumn{9}{l}{\textit{State Estimation}} \\
GCN         & 0.0166 & \textbf{0.00} & 0.0117 & \textbf{0.00} & 0.0502 & 0.49  & 0.0220 & 1.08 \\
GAT         & 0.0173 & \textbf{0.00} & 0.0116 & \textbf{0.00} & 0.0500 & 0.55  & \textbf{0.0218} & 0.67 \\
GPS         & 0.0179 & \textbf{0.00} & 0.0121 & 3.60  & 0.0601 & 3.70  & 0.0261 & 4.34 \\
GNS         & 0.0195 & 3.13  & \textbf{0.0017} & 14.15 & 0.0378 & 1.82  & 0.0402 & 6.64 \\
GNS-kit     & 0.0257 & 0.01  & 0.0062 & 17.02 & 0.0450 & 8.85  & 0.0302 & 10.01 \\
MxGPS-ind   & 0.0188 & \textbf{0.00} & 0.0045 & 2.44  & \textbf{0.0279} & 0.41  & 0.0355 & 0.02 \\
MxGPS-cross & \textbf{0.0147} & \textbf{0.00} & 0.0049 & 2.70  & 0.0434 & \textbf{0.00} & 0.0263 & 0.34 \\
\bottomrule
\end{tabular}%
}
\end{table*}

\begin{table*}[h]
\centering
\caption{In-distribution to zero-shot performance decay, averaged over all four zero-shot topologies. ID and ZS $\text{MAE}_{V_m}$ are 3-fold cross-validation means (p.u.); $\Delta$\,(\%) is the relative increase; ZS BVR is the average over the four zero-shot topologies. Best $\Delta$ per task in \textbf{bold}; $0\%$ ZS BVR entries also \textbf{bold}. $\dagger$: GPS, GAT, GCN exhibit in-distribution magnitude collapse on SSE (slope$_{V_m}<0.1$, $R^2<0.1$); their low $\Delta_\text{SSE}$ reflects a degraded in-distribution baseline, not genuine zero-shot robustness.}
\label{tab:degradation}
\footnotesize
\begin{tabular}{lcccccccc}
\toprule
 & \multicolumn{4}{c}{Power Flow} & \multicolumn{4}{c}{State Estimation} \\
\cmidrule(lr){2-5}\cmidrule(lr){6-9}
Model & ID $\text{MAE}_{V_m}$ & ZS $\text{MAE}_{V_m}$ & $\Delta$\,(\%) & ZS BVR\,(\%)
      & ID $\text{MAE}_{V_m}$ & ZS $\text{MAE}_{V_m}$ & $\Delta$\,(\%) & ZS BVR\,(\%) \\
\midrule
GCN         & 0.0043 & 0.0195 & +359\%          & 0.86          & 0.0201 & 0.0251 & +25\%$^\dagger$     & 0.39 \\
GAT         & 0.0027 & 0.0212 & +680\%          & 6.36          & 0.0202 & 0.0252 & +25\%$^\dagger$     & 0.30 \\
GPS         & 0.0023 & 0.0340 & +1364\%         & 13.95         & 0.0201 & 0.0290 & +44\%$^\dagger$     & 2.91 \\
GNS         & 0.0022 & 0.0200 & +790\%          & 11.41         & 0.0007 & 0.0248 & $+$3587\%           & 6.43 \\
GNS-kit     & 0.0064 & 0.0187 & +190\%          & 4.22          & 0.0045 & 0.0268 & +489\%              & 8.97 \\
MxGPS-ind   & 0.0218 & 0.0303 & \textbf{+39\%}  & \textbf{0.00} & 0.0039 & 0.0217 & +461\%              & \textbf{0.72} \\
MxGPS-cross & 0.0207 & 0.0326 & +57\%           & \textbf{0.00} & 0.0060 & 0.0223 & \textbf{+271\%}     & 0.76 \\
\bottomrule
\end{tabular}
\end{table*}

The accuracy inversion summarised in Table~\ref{tab:degradation} identifies the mechanism: models with the lowest in-distribution PF error degrade the most under topology shift, directly evidencing \emph{topology overfitting}. Averaged across all four zero-shot topologies, MxGPS degrades by only $39\%$ in PF $\text{MAE}_{V_m}$ ($0.0218 \to 0.0303$); by contrast, GNS-kit degrades by ${\approx}190\%$ ($0.0064 \to 0.0187$), GNS by ${\approx}790\%$, and GPS by over $14\times$, despite starting from substantially lower in-distribution error. Because MxGPS's in-distribution error is roughly $10\times$ higher than the specialists', we report absolute ID and ZS errors alongside $\Delta\%$ in Table~\ref{tab:degradation} and anchor the comparison on the 162/300-bus absolutes above, so that the relative framing cannot flatter the multi-task model. The inversion is direct evidence that single-task fine-tuning encodes topology-specific relational structure rather than transferable physics. For SSE, GPS, GAT, and GCN exhibit apparent low degradation ratios ($+25$--$44\%$), but this is an artifact of in-distribution magnitude collapse (Table~\ref{tab:sse_acc}): their in-distribution MAE of ${\approx}0.020$ already reflects mean-prediction behaviour, leaving no further performance to lose under distribution shift. Among models with meaningful in-distribution SSE accuracy, the MxGPS variants degrade least ($+271\%$ to $+461\%$, vs.\ GNS-kit $+489\%$ and GNS $+3587\%$). The multi-task joint training objective, which forces the shared encoder to represent operating constraints relevant to both SSE and PF simultaneously, acts as a strong implicit regulariser against topology overfitting.

A potential objection is that the $0\%$ BVR of MxGPS follows trivially from amplitude shrinkage toward the mean: the in-distribution PF slope is $0.48$--$0.58$, and the Naive $V_m{=}1$ predictor also posts $0\%$ BVR. This reading does not hold: post-hoc clipping of any baseline's predictions to $[V_m^{\min}, V_m^{\max}]$ trivially yields $0\%$ BVR but leaves the estimate elsewhere unchanged; clipping cannot produce the combination MxGPS achieves on the large zero-shot systems, where $0\%$ BVR coincides with the lowest (162-bus) or near-lowest (300-bus) zero-shot PF error.

The GPS vs.\ MxGPS comparison provides a controlled ablation of multi-task training, since GPS shares the same backbone architecture and hyperparameters as each individual MxGPS branch. The outcome is unambiguous: GPS achieves competitive in-distribution PF accuracy ($\text{MAE}_{V_m}=0.0023$) but collapses on SSE calibration ($\text{slope}_{V_m}=0.091$, $R^2=0.101$) and degrades by over $14\times$ under topology shift. MxGPS retains strong SSE calibration and achieves zero boundary violations on all four zero-shot PF topologies, at the cost of higher in-distribution PF error. The trade-off is a direct consequence of the shared encoder being pulled by two task-specific gradient signals simultaneously: it cannot overfit to PF-specific relational structure, preventing both the magnitude collapse that pure PF fine-tuning induces on SSE and the topology overfitting visible in the degradation table.

\textbf{Cross-branch attention ablation.}
Comparing the two variants, MxGPS-ind emerges as the stronger overall configuration: it degrades less on zero-shot PF ($+39\%$ vs.\ $+57\%$), is more accurate on in-distribution SSE ($0.0039$ vs.\ $0.0060$), and is smaller and faster ($1.6$\,M parameters, $0.13$\,ms/graph vs.\ $1.9$\,M, $0.27$\,ms). MxGPS-cross offers a modest in-distribution PF advantage ($0.0207$ vs.\ $0.0218$) and its only consistent zero-shot advantage is a lower SSE degradation ratio ($+271\%$ vs.\ $+461\%$); zero-shot SSE BVR is comparable between the variants ($0.76\%$ vs.\ $0.72\%$), both well below GNS-kit ($8.97\%$). We attribute the limited benefit of cross-branch attention at $K=2$ to redundancy: with only two branches, initialised from the same Phase-A checkpoint and already coupled through the shared encoder and the common physics loss, most cross-task information flows through shared parameters, so the explicit per-node exchange adds capacity, and a risk of diluting task specialisation, without opening genuinely new information pathways. We expect the mechanism to become more valuable as the task portfolio grows, where pairwise task relations are richer and the shared encoder alone cannot mediate all of them; analysing the learned cross-branch attention weights also offers a natural explainability probe of which tasks exchange information, which we leave to future work. On the strength of this comparison, we recommend MxGPS-ind as the default configuration at $K=2$.

\begin{table}[h]
\centering
\caption{Model efficiency: total parameters, average wall-clock time per training epoch, and inference latency per graph (GPU, single batch). Ranges cover PF and SSE tasks for single-task models; MxGPS values are for joint SSE$+$PF training. Best per column in \textbf{bold}.}
\label{tab:efficiency}
\footnotesize
\begin{tabular}{lrrr}
\toprule
Model & Parameters & Epoch (s) & Inference (ms/graph) \\
\midrule
GCN        & 69\,K   & 18--33  & \textbf{0.02--0.03} \\
GAT        & 219\,K  & 19--40  & 0.03--0.04 \\
GPS        & 820\,K  & 37--71  & 0.09--0.13 \\
GNS        & 6.1\,M  & 45--104 & 0.16--0.19 \\
GNS-kit    & 20.1\,M & 132--174 & 0.43--0.68 \\
\midrule
MxGPS-ind    & 1.6\,M & 614 & 0.13 \\
MxGPS-cross  & 1.9\,M & 734 & 0.27 \\
\bottomrule
\end{tabular}
\end{table}

\begin{figure}[!t]
\centering
\includegraphics[width=\columnwidth]{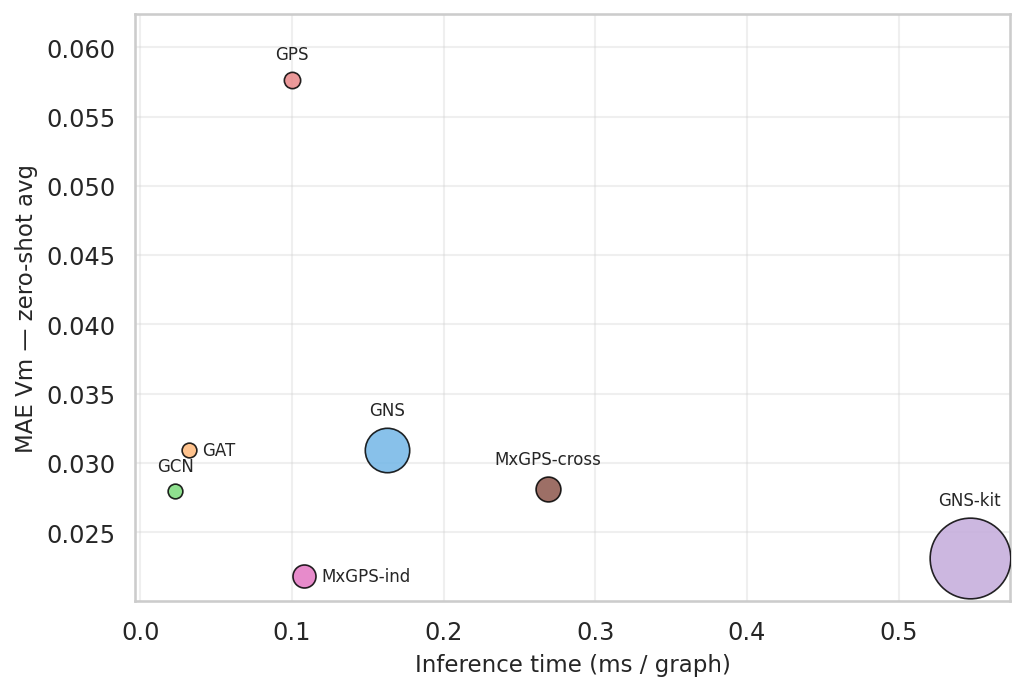}
\caption{Zero-shot PF accuracy--efficiency Pareto frontier: average $\text{MAE}_{V_m}$ across the four unseen topologies vs.\ inference time per graph (ms); bubble size proportional to parameter count. MxGPS-ind achieves the best zero-shot PF accuracy among models with $0\%$ BVR on all four topologies, at $12\times$ lower parameter cost than GNS-kit.}
\label{fig:efficiency}
\end{figure}

\subsection{Parameter Efficiency and Computational Cost}
\label{sec:efficiency}

Table~\ref{tab:efficiency} summarises parameter counts, per-epoch training time, and per-graph inference latency. For single-task baselines the ranges cover the PF and SSE fine-tuning phases (which use different batch sizes); for MxGPS the epoch time reflects joint SSE$+$PF training, with the linear probe stage training only 34\,K parameters for MxGPS-ind (300\,K for MxGPS-cross) before full unfreezing; single-task baselines train all parameters end-to-end throughout Phase~B.

MxGPS-ind (1.6\,M parameters, $12.6\times$ smaller than GNS-kit) achieves inference latency comparable to GPS ($0.13$\,ms/graph). Figure~\ref{fig:efficiency} shows the zero-shot PF accuracy--efficiency Pareto frontier: MxGPS-ind occupies the accuracy--efficiency sweet spot, matching or exceeding GNS-kit's zero-shot PF accuracy at $12\times$ lower parameter cost and $5\times$ lower inference latency. Critically, the efficiency advantage is not traded against zero-shot performance, a consequence of MxGPS achieving generalisation through the training objective rather than through scale.

\subsection{Limitations}
\label{sec:limitations}

The current experiments cover IEEE systems up to 300 buses; scaling to production grids requires replacing full $O(N^2)$ Transformer attention with a scalable approximation such as Exphormer. While both MxGPS variants achieve $0\%$ BVR on zero-shot PF, SSE zero-shot BVR remains non-zero on some topologies (up to $2.7\%$ for MxGPS-cross on case24). The multi-task portfolio currently covers SSE and PF only; extending to $K=3$ with N-1 contingency analysis requires only a new masking transform and prediction head, but the benefit of larger portfolios remains to be demonstrated empirically. Finally, the minimum training-portfolio size needed for reliable zero-shot generalisation remains an open question that warrants systematic study.

\section{Conclusions}
\label{sec:conclusions}

We have presented MxGPS, a multiplex graph transformer for multi-task power grid analysis, and identified \emph{topology overfitting}, the systematic failure of single-task GNN fine-tuning to generalise across grid topologies despite strong in-distribution accuracy, as the central failure mode that motivates its design. Models with the lowest in-distribution PF error degrade by $190\%$--$1400\%$ under topology shift, while MxGPS degrades by only $39\%$, maintains $0\%$ BVR on all four zero-shot PF topologies, and attains the lowest zero-shot PF error on the 162-bus system outright. The multi-task gradient signal also prevents the voltage magnitude collapse observed in single-task SSE fine-tuning, a failure mode that bare accuracy metrics conceal. Both results share a common mechanism: the shared encoder, forced to simultaneously satisfy complementary gradient signals from SSE and PF, cannot overfit to topology-specific or task-specific relational patterns, at a fraction of the parameter cost of the GridFM reference baseline.

Our ablation indicates that at $K=2$ the independent-branch configuration is the stronger default, with explicit cross-branch attention expected to become more valuable as the task portfolio grows. Future directions include extending the branch portfolio to optimal power flow and N-1 contingency analysis, where the value of cross-branch attention and its interpretability through learned attention weights can be assessed at larger $K$; replacing $O(N^2)$ full attention with a scalable approximation for production-scale grids; multi-seed replication of the cross-validation protocol; and per-topology adaptation of the shared encoder to close the remaining gap on in-distribution accuracy.

\section*{Acknowledgments}

The work presented is based on research conducted within the framework of the AI.grids initiative and by the Horizon Europe European Commission project EnerTEF (Grant Agreement No. 101172887). The authors would like to thank CRESYM and all the AI.grids consortium partners for their fruitful discussions, remarks and observations. The content of the paper is the sole responsibility of its authors and does not necessary reflect the views of the EC.


\bibliographystyle{ieeetr}
\bibliography{References}

\end{document}